\documentclass[a4paper]{styles/svproc}
\usepackage{cite}
\usepackage{amsmath,amssymb,amsfonts}
\usepackage{algorithmic}
\usepackage{tikz}
\usepackage{graphicx}
\usepackage{textcomp}
\usepackage{xcolor}
\usepackage{url}
\usepackage{bm}
\usepackage{subcaption}
\usepackage{caption}
\usepackage{multirow}
\usepackage{multicol}
\usepackage[bottom]{footmisc}
\begin{document}

\mainmatter              

\title{Design and Attitude Control of an Underwater Quadruped Robot}
\titlerunning{Underwater Quadruped Robot}   

\author{Davide Molinaroli \and Mohit Singh \and Kostas Alexis}
\authorrunning{Davide Molinaroli et al.} 


\institute{Norwegian University of Science and Technology, Trondheim 7034, Norway \email{davide.molinaroli@gmail.com}}

\maketitle   

\begin{abstract}
Legged robots are versatile on land, but their use in underwater environments remains limited. Extending quadruped locomotion to water enables amphibious mobility with applications in inspection, environmental monitoring and disaster response. This paper presents the design, modeling, and experimental validation of a reproducible underwater quadruped robot. The robot is built around custom waterproof motor housings machined from polyoxymethylene plastic, which use off-the-shelf O-rings and dynamic shaft seals. A simplified model is derived to describe the dynamics of this underwater legged system, capturing how drag forces on spherical end effectors transmit torque to the floating base. Building on this model, a closed-loop attitude controller is developed using an error formulation defined on the special orthogonal group SO(3). The controller is evaluated both in simulation and experimentally in a water tank, where the robot tracks desired orientation setpoints in roll, pitch and yaw.
\end{abstract}

\section{Introduction}\label{sec:introduction}
    \begin{figure}[htbp]
\vspace{-5mm}
\centerline{\includegraphics[width=\columnwidth]{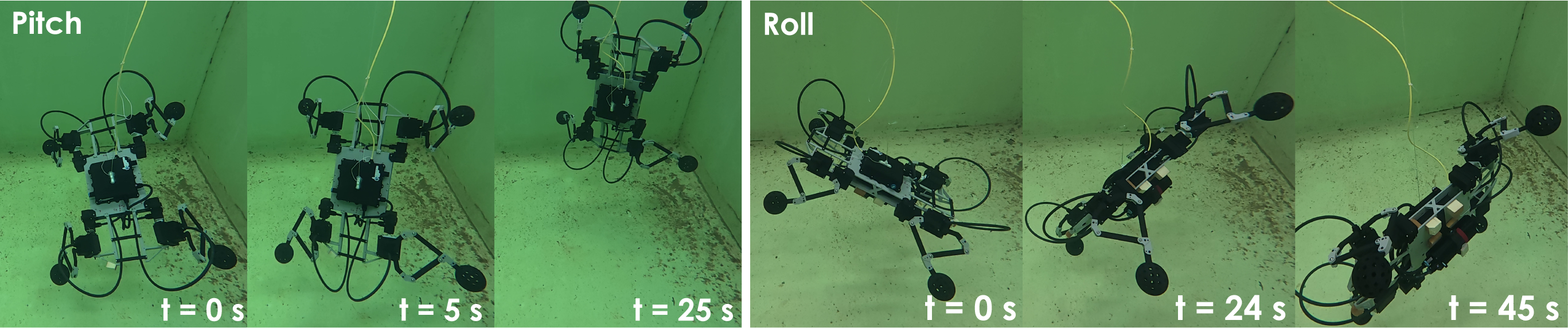}}
\vspace{-2ex}
\caption{The underwater quadruped robot performing pitch and roll maneuvers.}
\label{fig:pitch_roll_sequence}
\vspace{-5mm}
\end{figure}
Legged robots have shown strong capabilities in unstructured terrestrial environments, traversing rough terrain and navigating spaces that are inaccessible to wheeled or tracked vehicles~\cite{lee2020learning}. Extending their operational envelope to aquatic domains opens the possibility of amphibious operation, where a single platform can walk on land and swim underwater even without requiring separate propulsion mechanisms. Such robots could find use in underwater inspection and maintenance, environmental monitoring, and operations in partially flooded environments.

Despite this potential, achieving reliable underwater locomotion with a legged robot is challenging. From an engineering standpoint, the main difficulty lies in actuator design and realization - especially when it comes to reliably sealing low-cost off-the-shelf actuators not designed for underwater operation. From a control perspective, developing accurate hydrodynamic models for such systems is complex. A complete hydrodynamic treatment~\cite{fossen2011handbook} would require estimating added mass and drag coefficients for each body moving in the fluid. In this work, these quantities are not explicitly identified. Instead, a simplified model is derived that neglects added mass and captures the dominant drag forces. The results suggest that such a description is sufficient for attitude control, while not requiring extensive identification.

Overall, aiming to address the problem of underwater attitude control for a quadruped robot through drag-based actuation, this work contributes:
\begin{itemize}
\item The design and fabrication of an open-source, low-cost, and reproducible underwater quadruped robot using off-the-shelf servo motors, electronics, and sealing components, together with custom waterproof housings.
\item A simplified mathematical model, based on standard kinematic and dynamic principles, that captures how a legged underwater robot generates forces and torques via hydrodynamic drag and buoyancy on the end effectors.
\item A closed-loop attitude controller on $\mathrm{SO}(3)$ (Special Orthogonal group in 3D) with PID structure, anti-windup, and an allocation scheme that maps desired body torques into foot velocities, which are then integrated to foot positions and transformed to motor angle commands through inverse kinematics.
\item Experimental verification of the complete closed-loop system during attitude maneuvering in the water.
\end{itemize}

The remainder of the paper is organized as follows. Section~\ref{sec:relatedwork} overviews related work. Section~\ref{sec:robotdesign} outlines the robot design, followed by modeling in Section~\ref{sec:mathematicalmodeling}. The controller is detailed in Section~\ref{sec:controldesign}, and results are shown in Section~\ref{sec:experiments}, followed by discussion in Setion~\ref{sec:discussion}. Finally, conclusions are drawn in Section~\ref{sec:conclusion}.

\section{Related work}\label{sec:relatedwork}
In the literature, amphibious legged locomotion has been explored mainly through bio-inspired robots that mimic turtles~\cite{baines2022multi}, where locomotion is achieved by playing back recorded limb trajectories, or salamanders~\cite{ijspeert2020amphibious}, which use Central Pattern Generators. Recent work has started to investigate swimming with articulated legs for quadrupeds. In~\cite{patterson2026autonomous}, a sea turtle robot tracks transformed motion capture trajectories from green sea turtles using a PD joint space controller. The platform was deployed in a live reef environment to perform vision based animal following. In~\cite{qu2025amphibious}, hydrodynamic forces on a serial 2R planar leg are carefully modeled and paddling gaits are designed and analyzed. In~\cite{chen2024deep}, a beaver-like robot is studied and trained with reinforcement learning for forward swimming with pitch stabilization using planar legs. In~\cite{han2025learn}, an LSTM model approximating leg dynamics is trained on experimental data for planar motions and used to optimize gait parameters. In~\cite{chase2025design}, a quadruped  built around custom gearboxes and waterproof enclosures walks on the seafloor and swims to avoid obstacles based on a path planner. Related work on momentum-based reorientation has been studied for quadrupeds for in-flight attitude control~\cite{el2024flight, olsen2025olympus}, where leg motion redistributes angular momentum; nevertheless, this is distinct compared to the current work which in fact exploits hydrodynamic drag on submerged end-effectors.

\section{Robot Design}\label{sec:robotdesign}
\begin{figure}[t]
\centering
\begin{minipage}{0.48\columnwidth}
    \centering
    \includegraphics[width=\linewidth]{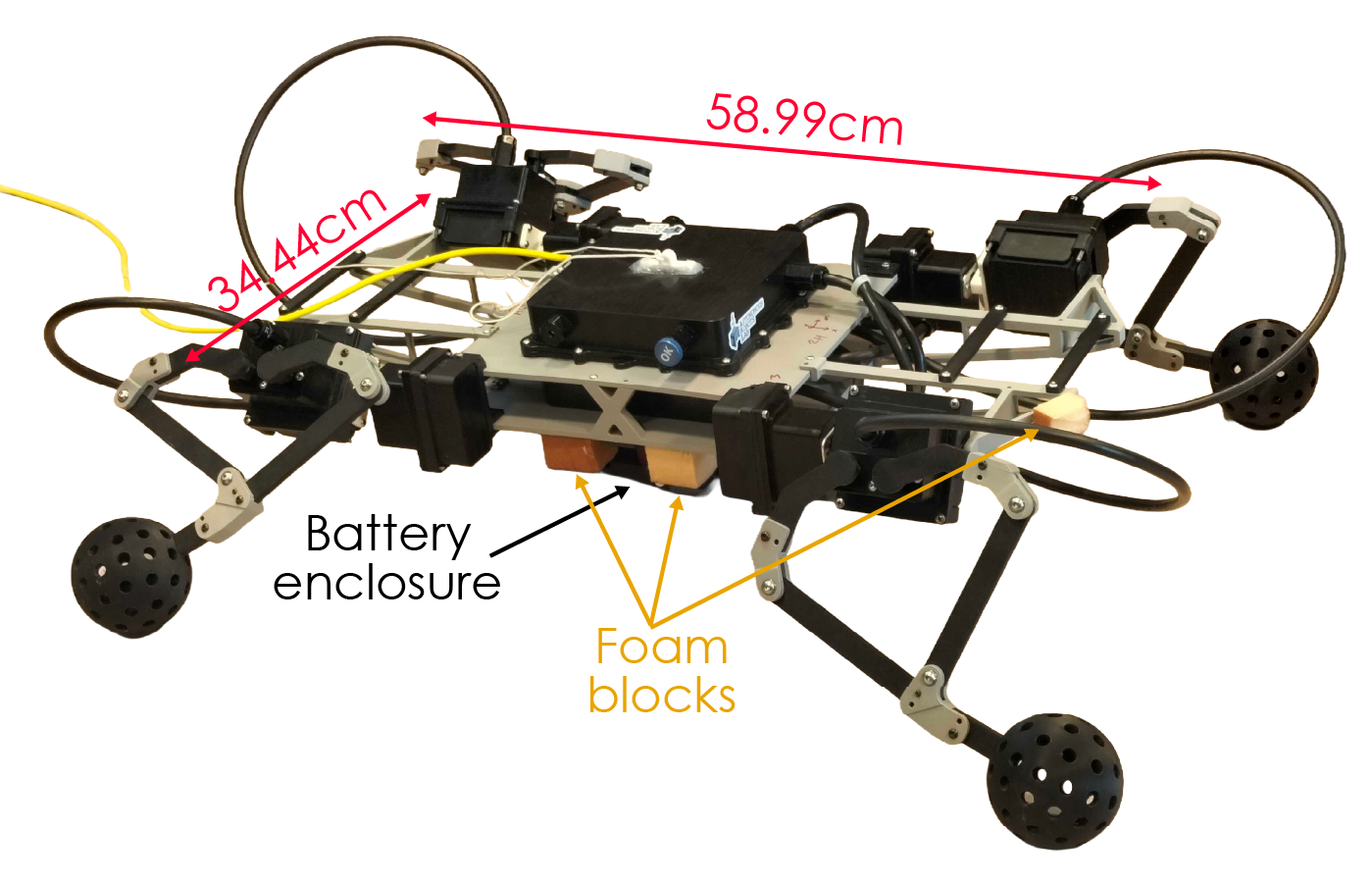}
    \caption{Assembly of the robot, highlighting the battery enclosure, foam blocks, and main body dimensions.}
    \label{fig:robot_real}
\end{minipage}
\hfill
\begin{minipage}{0.48\columnwidth}
    \centering
    \includegraphics[width=\linewidth]{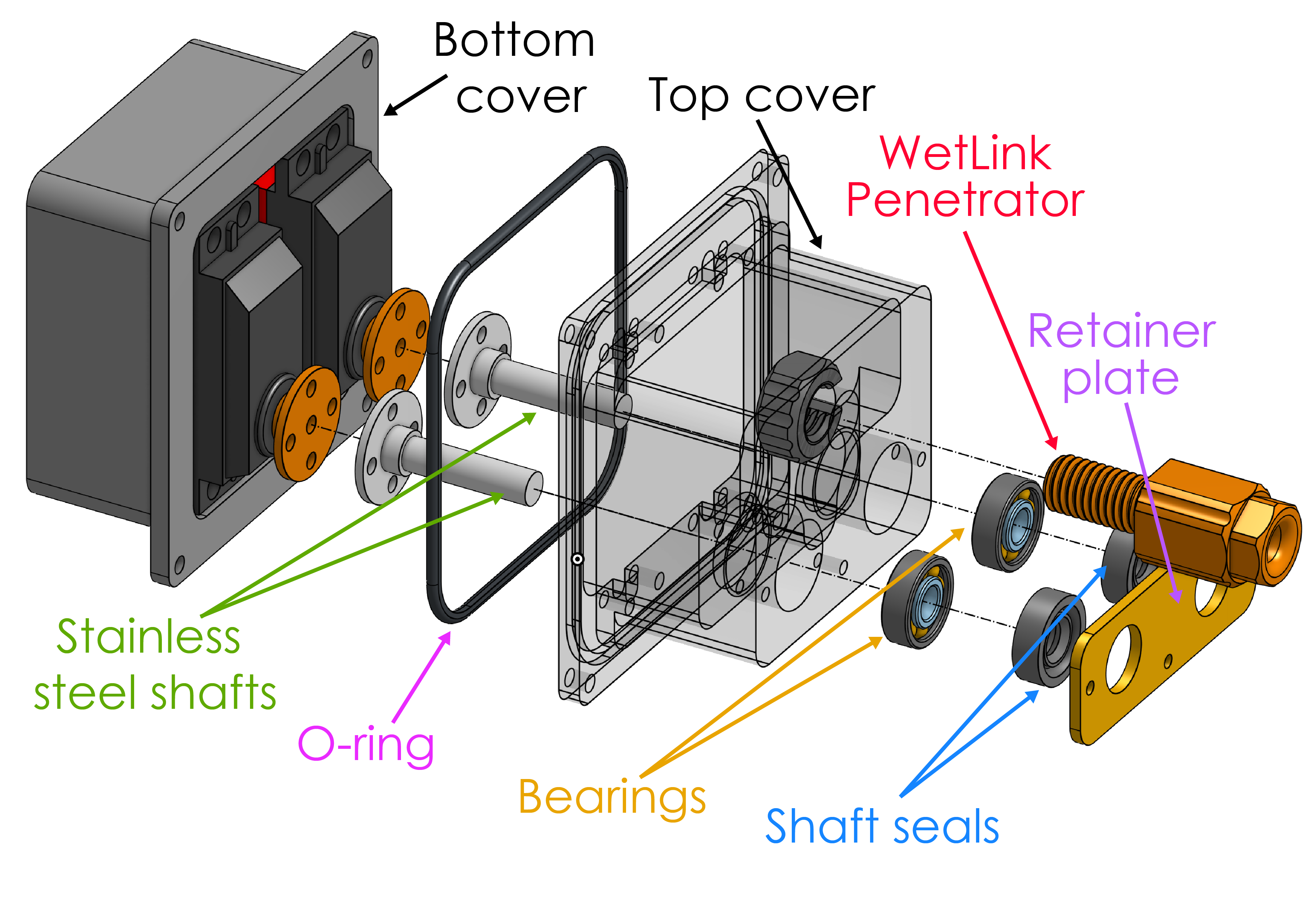}
    \caption{Exploded view of the waterproof casing for the lateral motors.}
    \label{fig:double_casing}
\end{minipage}
\vspace{-2mm}
\end{figure}

The robot, depicted in Fig.~\ref{fig:robot_real}, consists of a central waterproof electronics enclosure, four legs mirrored across the body and a cylindrical battery housing mounted beneath the main casing. Each leg is driven by three servo motors wrapped in waterproof enclosures: one controls the hip rotation about the longitudinal axis, while two lateral motors actuate a planar five-bar parallel mechanism that determines the leg's reach in the sagittal plane. The five-bar linkage allows two motors to be enclosed in a single casing, so that two sealed housings per leg could be used. Additionally, it enables load sharing between actuators and brings the leg's center of mass closer to the hip.

The approximate dimensions of the robot's body are 58.99~cm $\times$ 34.44~cm $\times$ 15.33~cm and its mass is  4.9~kg. Foam blocks are added below the robot to bring the center of buoyancy and center of mass as close as possible when the legs are symmetrically extended, while keeping the system slightly negatively buoyant to prevent it from floating.

CAD files are available at \url{https://ntnu-arl.github.io/underwater-quadruped/}.

\subsection{Waterproofing Strategy}

Waterproofing the joints and electronics is a central challenge in the mechanical design. A hard-casing approach was adopted, inspired by conventional ROV design, where every electronic component is enclosed in a rigid, sealed housing. Each motor casing is composed of two halves machined from POM (polyoxymethylene) plastic, bolted together with a commercially available O-ring squeezed between them to form a static seal. The O-ring grooves were dimensioned following~\cite{seals1992parker}. Each servo drives a custom stainless steel shaft that extends to the underwater environment and forms the revolute joint. A radial lip seal prevents water ingress during rotation, supported by a bearing that absorbs radial loads, preventing damage to the seal lip. Both are press fitted into the upper half of the casing, and a retainer plate keeps them in place to prevent loosening due to servo vibrations. Power and signal wires are routed via BlueRobotics WetLink Penetrators~\cite{BlueRobotics}, positioned to allow all three motors to rotate their full 180 degree range. The exploded view of one of the casings is shown in Fig.~\ref{fig:double_casing}. The sealing was validated through vacuum testing, followed by a two-hour submerged test at 1.5m of depth with servos continuously sweeping their full range. No water ingress was detected.

\begin{figure*}[t]
\centering
\includegraphics[width=0.95\textwidth]{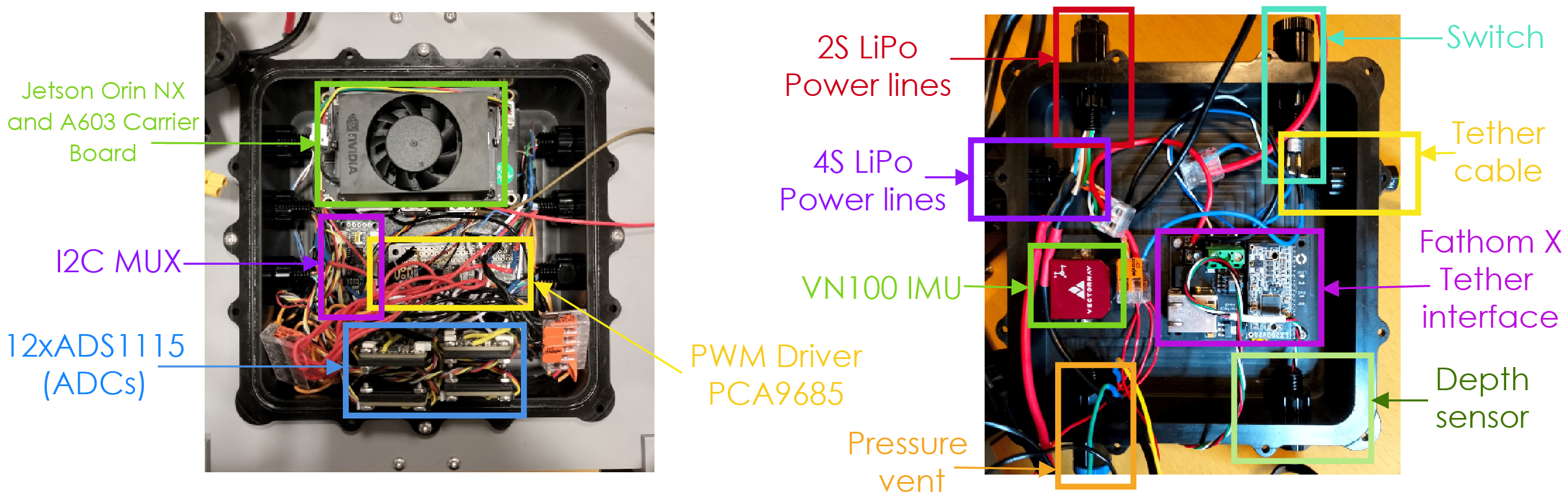}
\caption{Main electronics enclosure. The bottom part (left) contains the Jetson Orin NX and A603 board, I2C Mux, PWM driver and 12 ADCs. The top part
(right) contains power lines, tether communication boards and cables, IMU, pressure vent for vacuum testing, OBC switch and depth sensor.}
\label{fig:electronics}
\vspace{-1mm}
\end{figure*}
The main enclosure follows the same sealing principles and contains the onboard electronics (including a VectorNav VN-100 IMU) as shown in Fig.~\ref{fig:electronics}. A total of 14 WetLink Penetrators route motor cables, battery power, and tether cable to the main enclosure. Power is supplied by a 2S 3500~mAh LiPo battery for the motors and a 4S 550~mAh LiPo battery for the electronics, both housed in a BlueRobotics enclosure~\cite{BlueRobotics}. During the experiments described in Sect.~\ref{sec:experiments}, this setup provides approximately 35~minutes of continuous operation.

\subsection{Leg and End Effector Design}
The legs use a planar five-bar linkage with link lengths $a_1 = a_2 = 80$~mm and $a_3 = a_4 = 110$~mm and motor spacing of $c = 30$~mm, as seen in Fig.~\ref{fig:five_bar_kinematics}. The zero positions of the servos were chosen to maximize the workspace.

The end effector was chosen to be a sphere because its hydrodynamic behavior is isotropic: it produces no lift and its drag depends only on the paw speed and cross-sectional area~\cite{roos1971some}, thus simplifying the mathematical model. The sphere has a diameter of 7.5~cm, is hollow and presents evenly distributed holes, which increase drag without introducing asymmetric force patterns and also reduce buoyancy.

For actuation, the IB53BHU servo motors from AGF-RC were selected for all twelve joints. These are low-cost, off-the-shelf RC servos that provide 3.9~Nm of torque and 95~rpm. They provide an analog feedback signal, which is read by twelve ADS1115 ADCs and is useful for system identification and control.

\subsection{Software Architecture}

The onboard software runs on ROS1. The main control node operates at 100~Hz and reads ADC feedback,  computes desired end effector positions from the attitude controller, solves inverse kinematics and sends PWM commands to the servos. A separate node interfaces with the VN-100 IMU and publishes orientation quaternions and angular velocities.

\section{Mathematical Modeling}\label{sec:mathematicalmodeling}
\begin{figure*}[t]
\centering
\begin{subfigure}[b]{0.36\textwidth}
    \centering
    \includegraphics[height=95px]{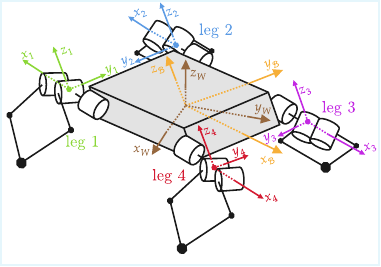}
    \caption{}
    \label{fig:robot_frames}
\end{subfigure}
\hfill
\begin{subfigure}[b]{0.25\textwidth}
    \centering
    \includegraphics[height=87.5px]{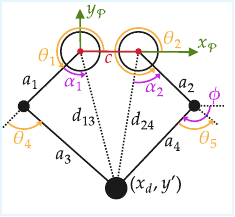}
    \caption{}
    \label{fig:five_bar_kinematics}
\end{subfigure}
\hfill
\begin{subfigure}[b]{0.32\textwidth}
    \centering
    \includegraphics[height=87.5px]{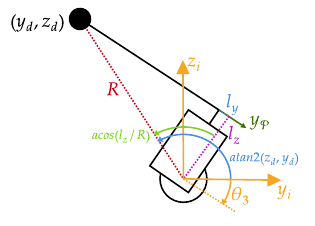}
    \caption{}
    \label{fig:leg_kinematics_hip}
\end{subfigure}

\caption{\textbf{(a)} Reference frames include world ($\mathcal{F}_W$), body ($\mathcal{F}_B$), and leg-specific ($\mathcal{F}_{L,i}$) systems. \textbf{(b)} The five-bar linkage uses actuated angles $\theta_1, \theta_2$ and passive angles $\theta_4, \theta_5, \phi$, with frame $\mathcal{F}_P$ at center of the innermost lateral motor. \textbf{(c)} Leg geometry shows hip motor angle $\theta_3$ and offsets $l_y, l_z$ from the origin of $\mathcal{F}_P$ to that of $\mathcal{F}_{L,i}$.}
\label{fig:kinematics_overview}
\end{figure*}
\par The underwater quadruped is modeled as a floating base with four kinematic chains. The model is built on the assumption that the dominant actuation mechanism is the drag force generated by each leg's spherical end effector as it moves through water. The dynamics of the legs are neglected. Only the end effector kinematics and the resulting drag and buoyancy forces/torques are considered.

\subsection{Reference Frames}

The reference frames of the model are shown in Fig.~\ref{fig:kinematics_overview}. The body frame $\mathcal{F}_B$ is centered on the robot's torso and shares its principal axes. Each leg $i = 1, ..., 4$ has a leg frame $\mathcal{F}_{L,i}$ whose origin is located at the intersection between the axes of the hip and innermost lateral motor. The four legs are arranged symmetrically: the body design is mirrored with respect to the $x_Bz_B$ and $y_Bz_B$ planes. For legs 1 and 2, which sit on the opposite side of the $y_Bz_B$ plane from legs 3 and 4, a 180-degree rotation about the $z$-axis is needed to align vectors from the leg frame into the body frame. Moreover, in forward and inverse kinematics, the $y$-axis of each leg frame is assumed to point inwards, as in Fig.~\ref{fig:leg_kinematics_hip}. Consequently, the $y$-coordinates of vectors expressed in leg frames 1 and 3 need to be inverted before being transformed into body frame quantities. These two operations are commutative and can be combined into a diagonal mirroring matrix $\mathbf{M}_i \in \mathbb{R}^{3 \times 3}$, defined per leg as:
\begin{equation}
\begin{aligned}
\mathbf{M}_1 &= \text{diag}(-1,\;1,\;1), &\quad \mathbf{M}_2 &= \text{diag}(-1,-1,\;1), \\
\mathbf{M}_3 &= \text{diag}(\;1,-1,\;1), &\quad \mathbf{M}_4 &= \text{diag}(\;1,\;1,\;1).
\end{aligned}
\label{eq:M_legs}
\end{equation}
The position of the $i$-th end effector in the body frame is then:
\begin{equation}
\mathbf{r}_{e,i}^B = \mathbf{o}_i^B + \mathbf{M}_i \, \mathbf{p}_{e,i}^{L}
\label{eq:ee_body}
\end{equation}
where $\mathbf{o}_i^B$ is the origin of the leg frame expressed in $\mathcal{F}_B$ and $\mathbf{p}_{e,i}^{L}$ is the end-effector position in the leg frame obtained from the forward kinematics. A two dimensional frame $\mathcal{F}_P$ is also attached to each plane of motion defined by the five-bar mechanism to solve closed chain kinematics locally.

\subsection{Leg Forward Kinematics}

The forward kinematics computes the end effector position in body frame given the motor angles $(\theta_1, \theta_2, \theta_3)$, where $\theta_1, \theta_2$ drive the five-bar mechanism and $\theta_3$ is the hip angle, defined as in Fig~\ref{fig:five_bar_kinematics} and \ref{fig:leg_kinematics_hip}. The derivation is identical for all legs in their own leg frame, so the index $i$ is omitted here. The position in the plane of the five-bar linkage is obtained following~\cite{demjen2023design}, exploiting the symmetry of the mechanism: the actuated links have equal length $a_1 = a_2$, the passive links have length $a_3 = a_4$, and $c$ is the distance between the motor axes. The offsets $l_y, l_z$ denote the origin of $\mathcal{F}_P$ relative to the hip motor's center of rotation. Defining:
\begin{align}
E &= 2a_3\bigl(c + a_1(\cos\theta_2 - \cos\theta_1)\bigr), \quad
F = 2a_1 a_3(\sin\theta_2 - \sin\theta_1), \\
G &= c^2 + 2a_1^2 + 2ca_1\cos\theta_2 - 2ca_1\cos\theta_1 \nonumber - 2a_1^2\cos(\theta_2 - \theta_1),
\label{eq:fk_EFG}
\end{align}
the passive angle $\phi$ of the outer kinematic chain is:
\begin{equation}
\phi = 2\,\text{atan2}\!\left(-F + \sqrt{E^2 + F^2 - G^2},\; G - E\right).
\label{eq:fk_phi}
\end{equation}
The end effector coordinates in $\mathcal{F}_P$ are then:
\begin{equation}
x_d = c + a_2\cos\theta_2 + a_4\cos\phi, \quad y' = a_2\sin\theta_2 + a_4\sin\phi.
\label{eq:fk_planar}
\end{equation}
The hip rotation $\theta_3$ about the $x$-axis maps the planar point $(x_d, y', 0)$ into the leg frame through:
\begin{equation}
\mathbf{p}_{e}^{L} = \begin{bmatrix}
    0 \\ t_y \\ t_z
\end{bmatrix} +  \begin{bmatrix} 1 & 0 & 0 \\ 0 & c_3 & -s_3 \\ 0 & s_3 & c_3 \end{bmatrix}\begin{bmatrix} x_d \\ y' \\ 0\end{bmatrix}
\label{eq:fk_T}
\end{equation}
where $c_3 = \cos\theta_3$, $s_3 = \sin\theta_3$, $t_y = l_y c_3 - l_z s_3$ and $t_z = l_y s_3 + l_z c_3$. The body frame position $\mathbf{r}_{e,i}^B$ is then obtained per leg through~\eqref{eq:ee_body}, applying the corresponding mirroring matrix and origin offset.

\subsection{Leg Inverse Kinematics}\label{sec:ik}
The inverse kinematics find the joint angles $(\theta_1, \theta_2, \theta_3, \theta_4, \theta_5)$ given a desired end effector position $\mathbf{r}_{e,i}^B$ in the body frame, first mapped to the leg frame by inverting~\eqref{eq:ee_body} as $\mathbf{p}_{e}^{L} = \mathbf{M}^{-1}_i(\mathbf{r}_{e,i}^B - \mathbf{o}_i^B) = (x_d, y_d, z_d)$, with the $y$ axis pointing inwards as in Fig.~\ref{fig:leg_kinematics_hip}. Once the end effector position is expressed in the leg frame, the procedure is identical for all legs, so the index $i$ is dropped. The hip angle is obtained as:
\begin{equation}
\theta_3 = \text{atan2}(z_d,\, y_d) - \arccos\left(\frac{l_z}{R}\right) - \frac{\pi}{2}
\label{eq:ik_theta3}
\end{equation}
where $R = \sqrt{y_d^2+z_d^2}$. The remaining angles are found by expressing the desired position onto the plane of the five-bar mechanism. Only the $y$ coordinate needs to be transformed, as $x_d$ is constant across the planar and leg frames:
\begin{equation}
    y' = y_d\cos\theta_3 + z_d\sin\theta_3 - l_y.
\end{equation} 
With reference to Fig.~\ref{fig:five_bar_kinematics}, the segments connecting the motors' centers to $(x_d,\, y')$ have length:
\begin{equation}
d_{13} = \sqrt{x_d^2 + y'^2}, \quad d_{24} = \sqrt{(x_d - c)^2 + y'^2},
\label{eq:ik_dists}
\end{equation}
and their angles between the physical links are:
\begin{equation}
\alpha_1 = \arccos\!\left(\frac{d_{13}^2 + a_1^2 - a_3^2}{2a_1 d_{13}}\right), \quad
\alpha_2 = \arccos\!\left(\frac{d_{24}^2 + a_2^2 - a_4^2}{2a_2 d_{24}}\right).\label{eq:ik_alphas}
\end{equation}
The actuated angles $(\theta_1, \, \theta_2)$ and the passive ones $(\theta_4, \, \theta_5)$ can be found as:
\begin{align}
\theta_1 &= \text{atan2}(y',\, x_d) - \alpha_1, &\quad
\theta_2 &= \text{atan2}(y',\, x_d - c) + \alpha_2,
\label{eq:ik_theta1245} \\
\theta_4 &= \arccos\!\left(\frac{d_{13}^2 - a_1^2 - a_3^2}{2\,a_1\,a_3}\right), &\quad
\theta_5 &= -\arccos\!\left(\frac{d_{24}^2 - a_2^2 - a_4^2}{2\,a_2\,a_4}\right). \nonumber
\end{align}

\subsection{Leg Translational Jacobian}
The translational Jacobian $\mathbf{J}_{P,i}^B \in \mathbb{R}^{3 \times 3}$ maps the motor velocities $(\dot\theta_1, \dot\theta_2, \dot\theta_3)$ to their contribution to the end effector velocity of leg $i$ in the body frame. Its first two columns capture the effect of the five-bar mechanism, and the third column accounts for the hip rotation.

Following~\cite{chavdarov2005kinematics}, the planar Jacobian of the five-bar mechanism with respect to the actuated angles $(\theta_1, \theta_2)$ is derived from the loop-closure constraint and also depends on the passive angles $(\theta_4, \theta_5)$. The derivation yields four scalar quantities $C_{11}$, $C_{21}$, $C_{12}$, $C_{22}$ that form the two columns of the $2 \times 2$ closed-chain Jacobian relating $(\dot\theta_1, \dot\theta_2)$ to the planar end effector velocity. The full translational Jacobian of the leg is then assembled as:
\begin{equation}
\mathbf{J}_{P,i}^B = \mathbf{M}_i \begin{bmatrix}\mathbf{R}_x(\theta_3)\begin{bmatrix}-C_{11}\\C_{21}\\0\end{bmatrix} & \ \mathbf{R}_x(\theta_3)\begin{bmatrix}-C_{12}\\C_{22}\\0\end{bmatrix} & \ \hat{\mathbf{x}} \times \mathbf{p}_{e,i}^L \end{bmatrix}
\label{eq:J_leg}
\end{equation}
where $\mathbf{R}_x(\theta_3)$ is the rotation matrix about the $x$-axis by $\theta_3$, $\hat{\mathbf{x}} = [1, 0, 0]^T$ is the hip rotation axis and $\mathbf{p}_{e,i}^L$ is the end effector position in the leg frame.

\subsection{End effector Velocity via Floating Base Kinematics}

The kinematics of each leg are described by the generalized coordinates $\mathbf{q}_i = [\mathbf{q}_b^T \quad \mathbf{q}_{j,i}^T]^T$, where $\mathbf{q}_b = [\mathbf{q}_{b_P}^T\quad  \mathbf{q}_{b_R}^T]^T$ collects the floating base position $\mathbf{q}_{b_P} \in \mathbb{R}^3$ and orientation quaternion $\mathbf{q}_{b_R} \in S^3$. The two lateral and hip joint angles of leg $i$ are, respectively, $\mathbf{q}_{j,i} = [\theta_{i,1} \quad \theta_{i,2} \quad \theta_{i,3}]^T$, with velocities $\dot{\mathbf{q}}_{j,i}$. The generalized velocity vector is $\mathbf{u}_i = [\mathbf{v}_B^{W\,T} \quad \bm{\omega}_B^{B\,T} \quad \dot{\mathbf{q}}_{j,i}^T]^T$, where $\mathbf{v}_B^W$ is the base's linear velocity in the world frame and $\bm{\omega}_B^B$ is the angular velocity in the body frame.

The linear velocity $\mathbf{v}_{e,i}^W$ of the $i$-th end effector in the world frame is obtained through the translational floating base Jacobian $\mathbf{J}_{P,i}^W$:
\begin{gather}
\mathbf{J}_{P,i}^W = \begin{bmatrix} \mathbf{I}_{3} & \ -\mathbf{R}_B^W [\mathbf{r}_{e,i}^B]_\times & \ \mathbf{R}_B^W \mathbf{J}_{P,i}^B(\mathbf{q}_{j,i}) \end{bmatrix}, 
\label{eq:ee_vel}\\ 
\mathbf{v}_{e,i}^W = \mathbf{J}_{P,i}^W(\mathbf{q}_i) \, \mathbf{u}_i.
\label{eq:fb_jac}
\end{gather}
Here, $ \mathbf{I}_{3}$ is the $3 \times3$ identity matrix, $\mathbf{R}_B^W$ is the rotation from the body frame to the world frame extracted from $\mathbf{q}_{b_R}$, $\mathbf{r}_{e,i}^B$ is the end effector position in the body frame as defined in~\eqref{eq:ee_body}, $[\cdot]_\times$ denotes the skew-symmetric operator, and $\mathbf{J}_{P,i}^B$ is the translational Jacobian defined in~\eqref{eq:J_leg}.

\subsection{Drag Force and Torque Transmission}
\label{sec:drag_torque}

Each spherical end effector moving with velocity $\mathbf{v}_{e,i}^W$ in still water experiences a drag force:
\begin{equation}
\mathbf{F}^W_{D,i} = -\tfrac{1}{2}\rho C_D A \|\mathbf{v}_{e,i}^W\| \, \mathbf{v}_{e,i}^W
\label{eq:drag}
\end{equation}
where $\rho$ is the fluid density, $C_D$ is the drag coefficient of the paw and $A$ is the cross-sectional area of the sphere.

By the principle of virtual work, the Cartesian force $\mathbf{F}^W_{D,i}$ applied at the end effector maps into generalized forces and torques through the transpose of the translational floating base Jacobian:
\begin{equation}
\bm{\tau}_{g,i} = \begin{bmatrix} \mathbf{F}_i^W \\ \bm{\tau}_i^B \\ \bm{\tau}_{j,i} \end{bmatrix} = (\mathbf{J}_{P,i}^W)^T \mathbf{F}^W_{D,i}.
\label{eq:virtual_work}
\end{equation}
Since leg dynamics are neglected, the joint torques $\bm{\tau}_{j,i}$ are discarded and only the world frame force $\mathbf{F}_i^W$ and body frame torque $\bm{\tau}_i^B$ contribute to the floating base dynamics. More explicitly, with $k= \tfrac{1}{2}\rho C_D A$ and $\mathbf{R}_W^B =\!(\mathbf{R}_B^W)^T$, leg $i$ exerts:
\begin{align}
\mathbf{F}_i^W &= -k \|\mathbf{v}_{e,i}^W\| \mathbf{v}_{e,i}^W
\label{eq:force_leg} \\
\bm{\tau}_i^B &= -k \|\mathbf{v}_{e,i}^W\| \, [\mathbf{r}_{e,i}^B]_\times \, (\mathbf{R}_W^B \, \mathbf{v}_{e,i}^W).
\label{eq:torque_leg}
\end{align}

\subsection{Paw Buoyancy Torque}

Each spherical paw has mass $m_s$ and volume $V_s$. When submerged, paw $i$ experiences a net vertical force in the world frame:
\begin{equation}
\mathbf{B}_i^W = (g\rho V_s - m_s g)\hat{\mathbf{z}}
\label{eq:buoy_force}
\end{equation}
where $g$ is the gravitational acceleration, $\rho$ is the fluid density and $\hat{\mathbf{z}} = [0, 0, 1]^T$. This force acts at the end effector position $\mathbf{r}_{e,i}^B$, producing a force and a torque on the body. The total buoyancy wrench acting on the body is:
\begin{align}
\mathbf{F}^W_{\text{buoyancy}} &= \sum_{i=1}^4 \mathbf{B}_i^W, &\quad
\bm{\tau}_{\text{buoyancy}}^B &= \sum_{i=1}^{4} [\mathbf{r}_{e,i}^B]_\times \left(\mathbf{R}_W^B \, \mathbf{B}_i^W\right).
\label{eq:buoy_torque}
\end{align}

\subsection{Inertia-Based Body Drag Model}

For simulation purposes, we adopt MuJoCo's inertia-based fluid model~\cite{MujocoFluidForces}, which approximates the body as a rectangular box whose half-dimensions are chosen to reproduce the diagonal inertia tensor. For a body of mass $m$ with principal moments $(I_{xx}, I_{yy}, I_{zz})$, the equivalent inertia box half-lengths are:
\begin{gather}
\begin{aligned}
r_x = \sqrt{\tfrac{3}{2m}(I_{yy}+I_{zz}-I_{xx})}, \qquad
r_y = \sqrt{\tfrac{3}{2m}(I_{zz}+I_{xx}-I_{yy})}
\end{aligned}
\nonumber \\
r_z = \sqrt{\tfrac{3}{2m}(I_{xx}+I_{yy}-I_{zz})}.
\label{eq:inertia_box}
\end{gather}
The drag on the body consists of a quadratic term and a viscous term. The $i$-th component of the quadratic drag force and torque, with $(i,j,k)$ a cyclic permutation of $(x,y,z)$, reads:
\begin{equation}
f_{D,i} = -2\rho \, r_j r_k \, |v_i^B| \, v_i^B, \qquad
g_{D,i} = -\tfrac{1}{2}\rho \, r_i(r_j^4 + r_k^4) \, |\omega_i^B| \, \omega_i^B.
\label{eq:quad}
\end{equation}
where $r_j r_k$ is proportional to the cross-sectional area of the box face normal to axis $i$, $v_i^B$ and $\omega_i^B$ are the $i$-coordinates of linear and angular velocity in body frame, and $r_i(r_j^4+r_k^4)$ represents a reference rotational inertia for the surface swept by a rotation about axis $i$.

The viscous term is linear in velocity and is computed from Stokes' law for an equivalent sphere of radius $r_{\text{eq}} = (r_x + r_y + r_z)/3$:
\begin{equation}
f_{V,i} = -6\beta\pi \, r_{\text{eq}} \, v_i^B, \qquad
g_{V,i} = -8\beta\pi \, r_{\text{eq}}^3 \, \omega_i^B.
\label{eq:visc}
\end{equation}
where $\beta$ is the dynamic viscosity of the fluid. Stacking the components of~\eqref{eq:quad} and~\eqref{eq:visc} into vectors $\mathbf{f}_D, \mathbf{g}_D, \mathbf{f}_V, \mathbf{g}_V \in \mathbb{R}^3$, the complete drag force is rotated to the world frame, while the torque remains in the body frame:
\begin{equation}
\mathbf{f}_{\text{drag}}^W = \mathbf{R}_B^W (\mathbf{f}_D + \mathbf{f}_V), \qquad \bm{\tau}_{\text{drag}}^B = \mathbf{g}_D + \mathbf{g}_V.
\label{eq:drag_total}
\end{equation}
This model neglects added mass and lift, but its coefficients are computed directly from the robot's geometric properties without experimental identification.

\subsection{Body Dynamics}

The floating base dynamics are described by the Newton-Euler equations, in which the contributions of all four legs~\eqref{eq:force_leg}-\eqref{eq:torque_leg}, body drag~\eqref{eq:drag_total} and buoyancy~\eqref{eq:buoy_torque} are summed as external forces and torques:
\begin{align}
m \dot{\mathbf{v}}_B^W = \sum_{i=1}^{4} \mathbf{F}_i^W + \mathbf{F}^W_{\text{buoyancy}} + \mathbf{f}_{\text{drag}}^W
\label{eq:trans}\\
\mathbf{I}\dot{\bm{\omega}}_B^B + \bm{\omega}_B^B \times (\mathbf{I}\bm{\omega}_B^B) = \sum_{i=1}^{4} \bm{\tau}_i^B + \bm{\tau}_{\text{buoyancy}}^B + \bm{\tau}_{\text{drag}}^B
\label{eq:rot}
\end{align}
where $m$ is the body mass and $\mathbf{I}$ is the body inertia tensor. The
gravity and buoyancy forces on the torso are not modeled: the real robot is slightly negatively buoyant and during experiments it is suspended from a rope, so these forces are balanced by the rope tension and can be neglected. It is further assumed that the center of mass and center of volume of the torso coincide, so that the torso's buoyancy produces no torque. The commanded joint angles $\mathbf{q}_{j,i}$ are the inputs to this model and their velocities $\dot{\mathbf{q}}_{j,i}$ are obtained through finite differences.

\section{Attitude Controller Design}\label{sec:controldesign}
The controller drives the robot from its current orientation to a desired one by computing a total drag torque that the legs should produce on the body. It consists of three layers: an error computation on $\mathrm{SO}(3)$, a PID control law structure that produces a torque command and an allocation scheme that converts this torque into foot velocity commands. The controller acts on the rotational dynamics~\eqref{eq:rot} and treats the buoyancy torque as an estimated disturbance that is compensated through feedforward.

\subsection{Orientation Error}
Let $\mathbf{R}_d \in \text{SO}(3)$ be the desired rotation matrix and $\mathbf{R}_B^W \in \text{SO}(3)$ be the current rotation extracted from the IMU measurement. Following~\cite{lee2010quadrotor}, the orientation error vector $\mathbf{e}_R \in \mathbb{R}^3$ is extracted from the skew-symmetric part of the relative rotation $\mathbf{R}_d^T \mathbf{R}_B^W$:
\begin{equation}
[\mathbf{e}_R]_\times = \frac{1}{2}\left(\mathbf{R}_d^T \mathbf{R}_B^W - (\mathbf{R}_B^W)^T \mathbf{R}_d \right)
\label{eq:error}
\end{equation}
and recovered via the vee map. This formulation avoids the singularities of Euler angle representations and yields an error signal that vanishes if and only if $\mathbf{R}_B^W = \mathbf{R}_d$.

\subsection{Control Law}
The total commanded torque in the body frame is:
\begin{align}
\bm{\tau}_{\text{cmd}} &= -K_p \mathbf{e}_R - K_i \int_0^t (\mathbf{e}_R + K_{\text{aw}} \bm{\Delta}_{\text{sat}}) \, d\tau - K_d \bm{\omega}_B^B \nonumber \\
&\quad + \bm{\omega}_B^B \times (\mathbf{I}\bm{\omega}_B^B) - \bm{\tau}_{\text{buoyancy}}^B
\label{eq:pid}
\end{align}
where $K_p$, $K_i$, $K_d$ are scalar gains, the gyroscopic coupling $\bm{\omega}_B^B \times \mathbf{I}\bm{\omega}_B^B$ and the buoyancy torque are compensated as feedforward terms. The control law stabilizes the attitude error dynamics of a rigid body where torque can be instantaneously controlled~\cite{lee2010quadrotor}. Although the robot is a multi-body system, for the purposes of control, it is assumed that the leg dynamics are sufficiently fast compared to those of the torso such that they can be neglected and treated as a virtually instantaneous actuator.

The commanded torque magnitude is clamped to $\tau_{\max} = 2$~Nm:
\begin{equation}
\bm{\tau}_{\text{sat}} = \hat{\bm{\tau}}_{\text{cmd}} \min(\|\bm{\tau}_{\text{cmd}}\|, \tau_{\max}),
\label{eq:sat}
\end{equation}
where the hat denotes unit vectors. The value of $\tau_{\max}$ was estimated by evaluating~\eqref{eq:torque_leg} using IMU angular velocity and joint velocities filtered from servo feedback during preliminary trials run without torque clamping, thus reflects the maximum torque the four legs can collectively produce. The saturation difference $\bm{\Delta}_{\text{sat}} = \bm{\tau}_{\text{cmd}} - \bm{\tau}_{\text{sat}}$ is fed back through the gain $K_{\text{aw}}$ as an anti-windup mechanism that slows the accumulation in the integrator during saturation.

\subsection{Torque Allocation and Foot Velocity}

The saturated torque is distributed equally among the four legs: $\bm{\tau}_i^B = \bm{\tau}_{\text{sat}} / 4$. The drag force direction is computed to be perpendicular to the paw position $\mathbf{r}_{e,i}^B$ and $\bm{\tau}_i$ as $\hat{\mathbf{f}}_i = \hat{\bm{\tau}}_i \times \hat{\mathbf{r}}_i$. Under the orthogonality assumption and from~\eqref{eq:torque_leg}, the required foot speed is:
\begin{equation}
v_i = \sqrt{\frac{\|\bm{\tau}^B_i\|}{k \|\mathbf{r}_{e,i}^B\|}}
\label{eq:foot_speed}
\end{equation}
The foot velocity command including compensation for body rotation is:
\begin{equation}
\mathbf{v}_i^{\text{cmd}} = -\hat{\mathbf{f}}_i v_i - \bm{\omega}_B^B \times \mathbf{r}_{e,i}^B.
\label{eq:foot_vel}
\end{equation}

\subsection{Power and Recovery Stroke Cycle}

The legs are grouped into two diagonal pairs, $(1,2)$ and $(3,4)$, which alternate between power and recovery phases. The control law is stabilizing under continuous torque generation, but two factors prevent this on the real system: each leg operates in a finite workspace and must periodically retract, and the position-controlled servos have limited bandwidth. Two measures are adopted to mitigate these effects. First, pair $(1,2)$ starts in a half-duration recovery phase in an attempt to generate a more continuous torque signal by introducing an initial time delay between the power phase of the two leg pairs. Second, the velocity command $\mathbf{v}_i^{\text{cmd}}$ is computed once at the start of the phase and held constant for its duration, giving the position controlled servos time to reach the corresponding joint velocities. The foot velocity is integrated with Euler's method to obtain a body frame position, starting from a nominal position near the workspace center:
\begin{equation}
\mathbf{r}_{e,i}^B(t + \Delta t) = \mathbf{r}_{e,i}^B(t) + \mathbf{v}_i^{\text{cmd}}\,\Delta t.
\label{eq:foot_integration}
\end{equation}
The resulting position is mapped to joint angles using the inverse kinematics of Sect.~\ref{sec:ik}. The power phase terminates when the inverse kinematics solver fails, indicating a workspace boundary, or after a maximum duration of $t^\text{max}_\text{pow} = 0.5$~s

The trajectory is recorded and retraced in reverse during the recovery phase, parametrized in arc length and linearly interpolated with a cubic time profile ensuring zero velocity at the endpoints, over a duration $t_{\text{rec}} \ge 2 t^\text{max}_\text{pow}$.

\subsection{Simulation Validation}

\begin{figure}[t]
\centerline{\includegraphics[width=0.75\columnwidth]{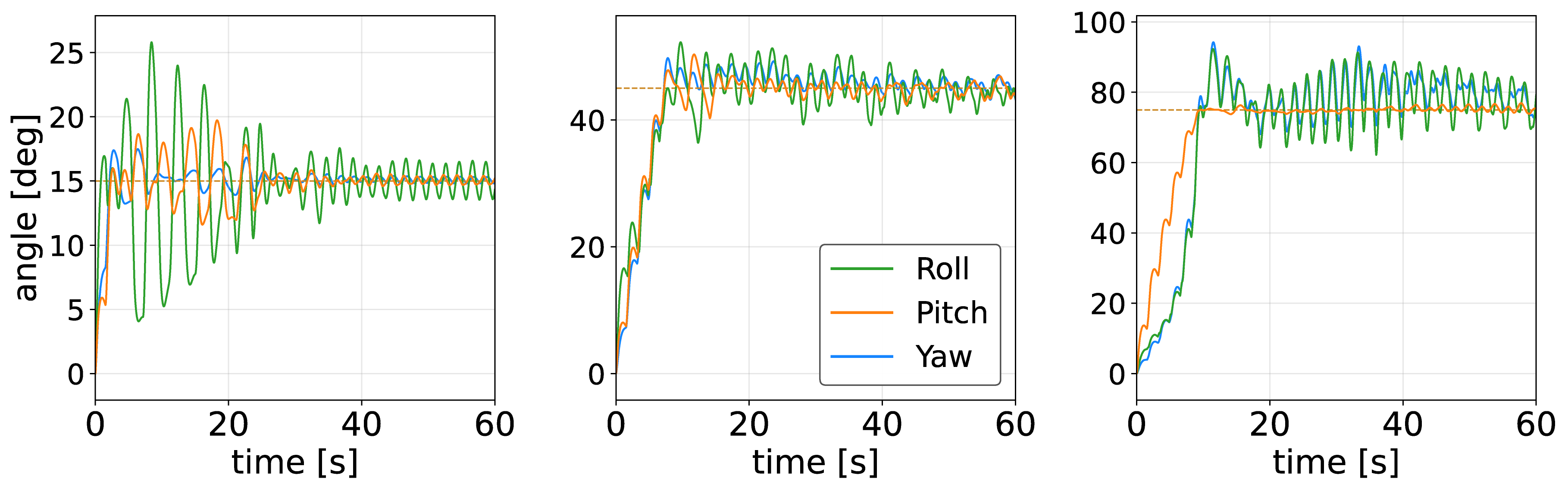}}
\caption{Simulation results of the closed-loop attitude controller on three trials. Step setpoints of amplitudes from 15 to 75 degrees are applied to all three axes simultaneously with increments of 30 degrees between consecutive trials.}
\label{fig:sim}
\end{figure}

Before deployment on the real robot, the controller was evaluated in simulation using the model from ~\eqref{eq:trans}-\eqref{eq:rot}, integrated with a fourth-order Runge-Kutta scheme with step size $\Delta t = 0.01$~s. The robot was commanded to track constant orientation setpoints from 15 to 75 degrees in 30-degree increments, applied simultaneously to roll, pitch, and yaw. The gains were set to $K_p = 5$, $K_i = 0.15$, $K_d = 1$, and $K_{\text{aw}} = 0.25$, with $t_{\text{rec}} = 1.5$~s. The drag coefficient of the paw was set to $C_D = 0.64$, estimated using Computational Fluid Dynamics with the SIMPLE~\cite{ferziger2002computational} algorithm , in a cube of volume $1.25\text{m}^3$, at Reynolds number $\text{Re}=70000$. The simulated controller tracks all setpoints with steady-state oscillations arising from the power-recovery stroke mechanism, as seen in Fig.\ref{fig:sim}.

\section{Experiments}\label{sec:experiments}
\setlength{\tabcolsep}{3pt}
\begin{table}[t]
\centering
\scriptsize
\caption{Regulation performance across experiments.}
\label{tab:tracking_metrics}
\begin{tabular}{ccccccc|ccccc}
\hline
 Set & Cmd & $\theta_{\text{init}}$ & $t_r$ & $t_s$ & $e_{\mathrm{ss}}$ & $M_p$ & $\psi_{\text{init}}$ & $t_r$ & $t_s$ & $e_{\mathrm{ss}}$ & $M_p$\\
 & [deg] & [deg] & [s] & [s] & [deg] & [\%] & [deg] & [s] & [s] & [deg] & [\%]\\
\hline
\multirow{6}{*}{Roll}
  & 15 & 80.89  & 23.34 & 51.64  & -0.45    & 58.0 & -78.02 & 12.41 & 25.67 & 0.37 & 13.9 \\
  & 30 & 50.86  & 15.85 & 34.06  & 1.48     & 26.1 & -35.49 & 5.38 & 18.16 & -1.14 & 11.5 \\
  & 45 & 132.86 & 39.91 & 61.20  & -0.51    & 28.3 & 121.83 & 21.03 & 46.36 & -0.54 & 14.8 \\
  & 60 & 84.54  & 35.61 & 64.94  & -0.83    & 17.9 & 71.15 & 10.82 & 37.15 & 1.09 & 15.5 \\
  & 75 & 115.21 & 20.45 & --     & -2.65    & 10.9 & 82.91 & 17.74 & 45.88 & -3.38 & 9.3 \\
  & 90 & 158.31 & 33.41 & 128.56 & 2.93     & 12.4 & 123.32 & 15.27 & 85.22 & 1.11 & 16.7 \\
\hline
\multirow{6}{*}{Pitch}
  & 15  & 9.35  & 2.19  & 2.91  & -0.32 & 32.4 & 1.54 & 0.81 & 0.00 & 0.32 & 2.2  \\
  & 30  & 64.58 & 11.94 & 18.91 & 0.08  & 19.5 & 61.38 & 8.81 & 23.85 & -0.38 & 9.1  \\
  & 45  & 64.68 & 16.86 & 45.58 & 0.44  & 14.0 & 56.76 & 7.66 & 25.66 & -0.87 & 10.2 \\
  & 60  & 79.15 & 21.21 & 42.16 & 1.66  & 15.0 & -54.06 & 8.16 & 10.99 & -0.36 & 4.0  \\
  & 75  & 77.72 & 45.92 & 74.83 & -0.05 & 3.1  & -43.40 & 5.82 & 9.82 & -0.12 & 4.8  \\
  & 90  & 76.97 & 22.24 & 53.77 & -4.82 & 0.0    & 58.46 & 6.75 & -- & 1.04 & -- \\
\hline
\multirow{3}{*}{\shortstack{Roll\\+Pitch}}
  & 30 & 19.59 & 8.04 (R)  & 27.53 (R) & 0.56 (P)  & 25.2 (R) & -0.54  & 10.36 & 0.00 & 0.45 & 3.7 \\
  & 45 & 51.01 & 53.06 (R) & 52.19 (R) & 0.58 (P)  & 15.1 (R) & -57.54 & 5.16 & 48.80 & 1.41 & 3.7 \\
  & 60 & 45.96 & 15.22 (R) & 38.76 (R) & 0.67 (R)  & 6.3 (R) & -38.35  & 10.79 & 36.00 & 1.16 & 4.1  \\
\hline
\end{tabular}
\end{table}

\subsection{Experimental Setup}

We conduct three sets of experiments to evaluate the attitude controller. In the first set, six roll setpoints are commanded, ranging from 15 to 90 degrees in 15-degree increments, with pitch and yaw references set to zero. In the second set, the same six setpoints are commanded on the pitch angle, with roll and yaw references set to zero. In the third set, three experiments are run in which roll and pitch are simultaneously commanded to 30, 45, and 60 degrees, with yaw reference set to zero. The yaw angle always starts from a different initial condition in the range from -78 to +123 degrees, so that yaw regulation to zero is always part of the control task.

The robot is suspended from a floating foam block connected to a fixed gantry above the water, using a hook attached to the top of the electronics enclosure. This constrains the robot to a fixed depth while reducing the torque disturbance that the rope tension exerts on the body during reorientation. 

The attitude controller was deployed with gains $K_p = 10$, $K_i = 3$, $K_d = 0.5$, $K_{\text{aw}} = 0.25$, and $t_{\text{rec}} = 1.5$~s. The gains were re-tuned relative to simulation due to differences between the model and the real system. In particular: the centers of mass and volume do not coincide exactly, introducing unmodeled restoring torques; the servos have finite bandwidth and do not reach the commanded velocities instantaneously; and added mass effects and leg dynamics were neglected.

Table~\ref{tab:tracking_metrics} summarizes the regulation performance across all experiments. Because the power-recovery cycle introduces oscillations in the robot's orientation, all signals are smoothed with a centered 3~s moving average before computing the metrics. Thus, the metrics reflect the average orientation that the robot sustains across stroke cycles rather than the instantaneous oscillations inherent to the control strategy. We also report the initial geodesic error $\theta_{\text{init}} = \arccos(\frac{\text{tr}(\mathbf{R}_d^T\mathbf{R}_B^W) - 1}{2})\big|_{t=0}$, which measures the total displacement the controller should correct along the geodesic. The rise time $t_r$ is the interval from 10\% to 90\% of the setpoint, the settling time $t_s$ is the earliest instant after which the error remains within $\pm6$ degrees for at least 30~s, $e_\mathrm{ss}$ is defined as the mean error over the last 10\% of the trial and $M_p$ is the percentage overshoot relative to the setpoint. For the coupled roll and pitch experiments, the table reports the worst-performing axis, indicated by (R) for roll or (P) for pitch. The yaw offset $\psi_\text{init}$ that must be driven to zero and yaw performance metrics are also shown.

\subsection{Roll Regulation}
\begin{figure}[t]

\centering
\begin{subfigure}[b]{0.49\columnwidth}
    \centering
    \includegraphics[width=\textwidth]{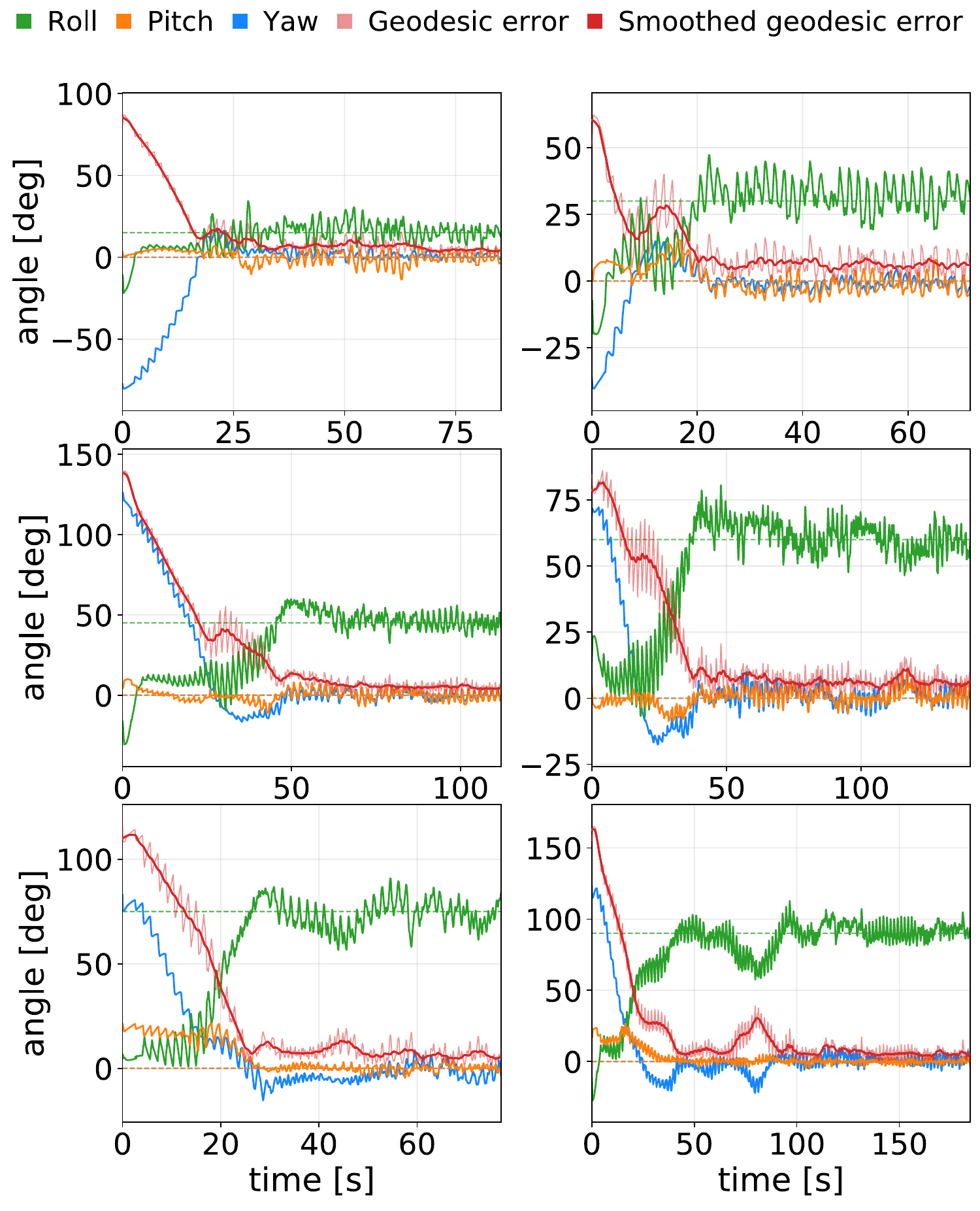}
    \caption{}
    \label{fig:experiments_roll}
\end{subfigure}
\hfill
\begin{subfigure}[b]{0.49\columnwidth}
    \centering
    \includegraphics[width=\textwidth]{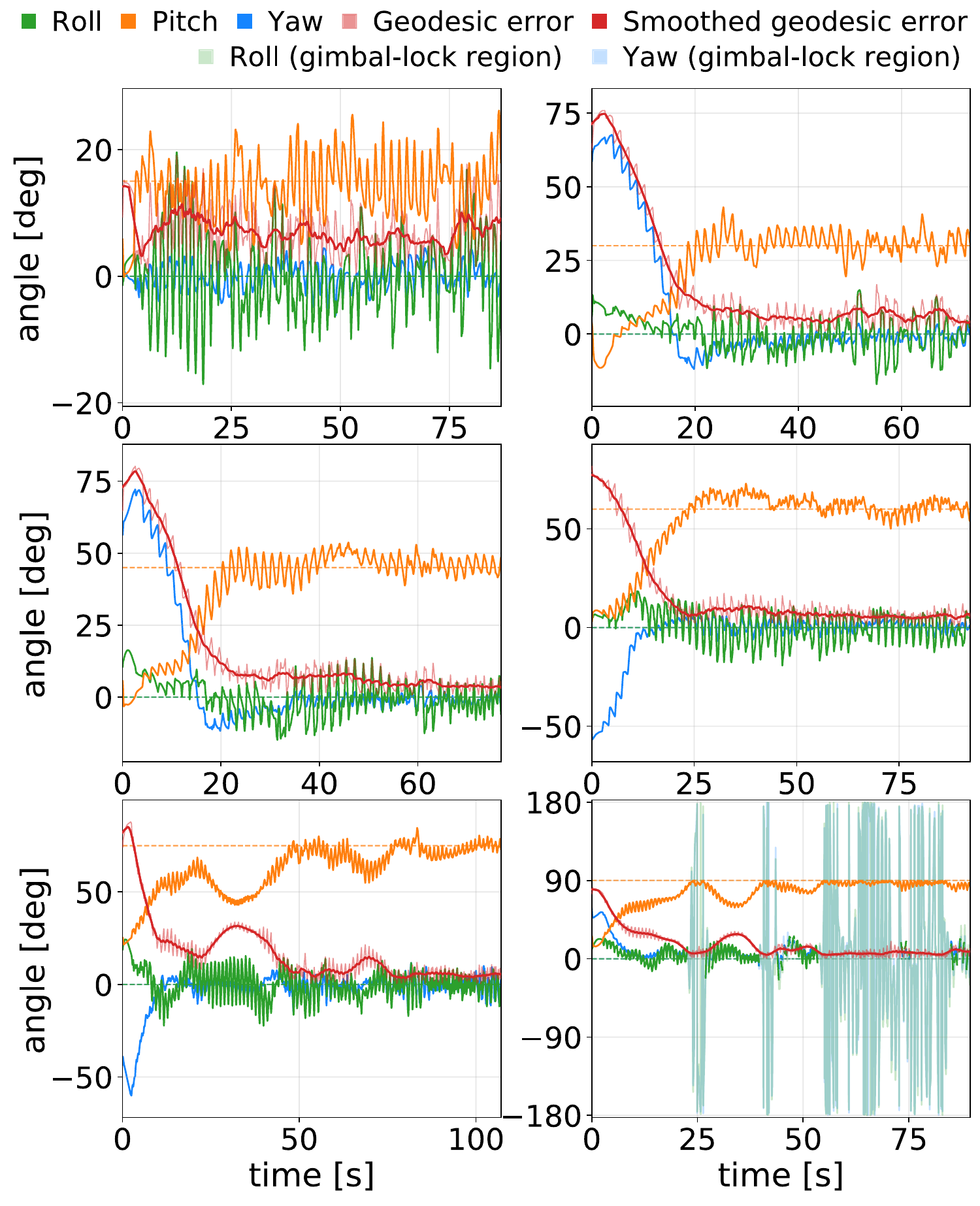}
    \caption{}
    \label{fig:experiments_pitch}
\end{subfigure}
\caption{Closed-loop attitude regulation experiments. \textbf{(a)} Roll setpoints from 15 to 90 degrees. \textbf{(b)} Pitch setpoints from 15 to 90 degrees. The controller corrects yaw from different initial conditions. At the 90-degree pitch setpoint, roll and yaw traces exhibit apparent jumps due to gimbal lock in the Euler angle representation. The geodesic error is the angle of the axis-angle decomposition of $\mathbf{R}_d^T \mathbf{R}_B^W$, plotted both raw and after a 3~s centered moving average.}
\label{fig:experiments}
\end{figure}

Roll setpoints from 15 to 90 degrees in 15-degree increments were commanded, with pitch and yaw held at zero. Results are shown in Fig.~\ref{fig:experiments_roll}. Settling times generally grow with the commanded amplitude, the exception being the 15-degree case where overshoot drives the smoothed roll signal outside the $\pm6$ degree tolerance band near $t=50$~s. At 75 degrees the smoothed signal does not remain within $\pm6$ degrees for 30 consecutive seconds, so no settling time is reported. Rise times do not follow the same pattern, as the rate at which each setpoint is first reached depends on the particular trajectory that the attitude dynamics take in $\mathrm{SO}(3)$.

\subsection{Pitch Regulation}
Pitch setpoints from 15 to 90 degrees in 15-degree increments were commanded, with roll and yaw held at zero. Results are shown in Fig.~\ref{fig:experiments_pitch}. In the first four experiments, settling and rise times grow with the amplitude of the setpoint. In the 75- and 90-degree experiments, the recovery phase duration was reduced to $t_{\text{rec}} = 1.0$~s to achieve stabilization. In these two experiments it is observed that, after a steep initial rise, the robot temporarily loses pitch. This occurs when the legs enter a relative phasing that is inefficient for torque generation. Workspace asymmetry then terminates the power strokes at different times across legs, breaking this phasing and driving the legs into a more favorable synchronization.

\subsection{Coupled Roll and Pitch Regulation}

\begin{figure}[t]
\centerline{\includegraphics[width=0.75\columnwidth]{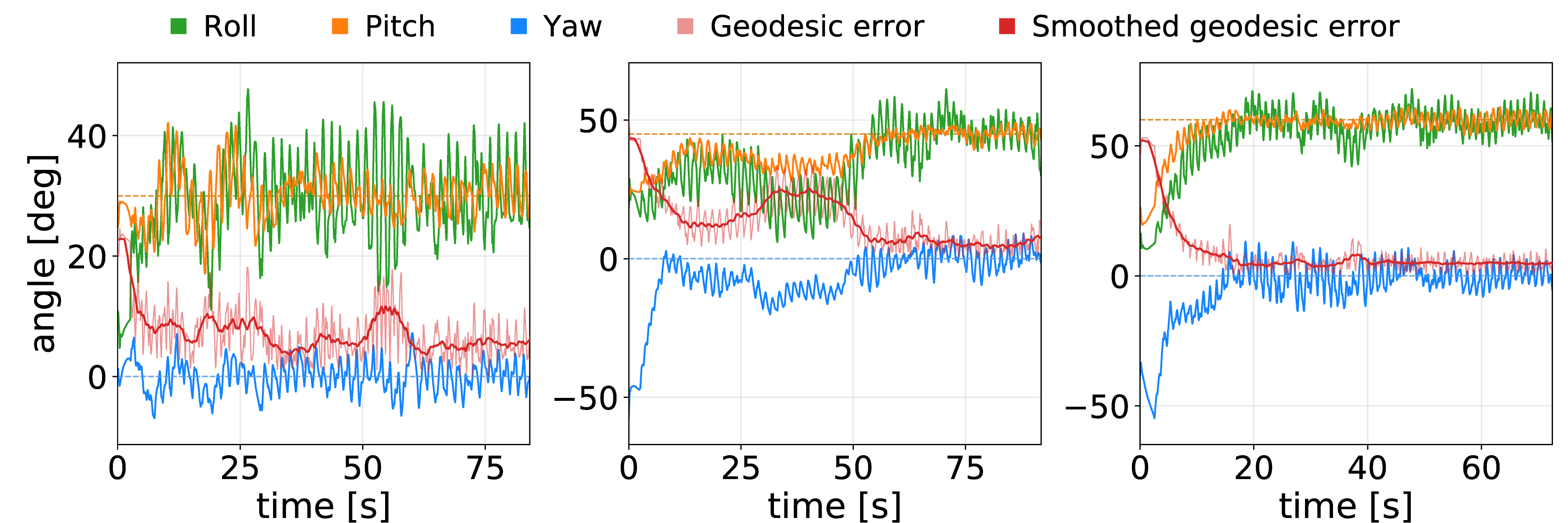}}
\caption{Regulation to coupled roll and pitch angles, while correcting yaw from different initial conditions. The geodesic error is the angle of the axis-angle decomposition of $\mathbf{R}_d^T \mathbf{R}_B^W$, plotted both raw and after a 3~s centered moving average.}
\label{fig:experiments_roll_and_pitch}
\end{figure}

In the coupled experiments, roll and pitch are commanded simultaneously to 30, 45 and 60 degrees. Results are shown in Fig.~\ref{fig:experiments_roll_and_pitch}. The robot stabilizes around all three setpoints. The 60-degree case required setting $t_{\text{rec}} = 1.0$~s, which yields a shorter rise and settling time than the 45-degree case.

\section{Discussion}\label{sec:discussion}
The results confirm that drag-based actuation through spherical end effectors can achieve attitude control of an underwater quadruped. The simplified model is sufficient to design a controller that transfers to the real system, though modeling servo and leg dynamics and hydrodynamic effects such as added mass and restoring forces could reduce the discrepancy between simulation and hardware.

Performance is primarily limited by speed: the lack of leg phase-coordination can cause simultaneous recovery phases, resulting in intermittent torque gaps and longer settling times. Additionally, the legs retract to a fixed position near the workspace center during recovery, so the power stroke covers roughly half of the available range. Improving upon this can increase the agility of the controller.

\section{Conclusion}\label{sec:conclusion}
This paper presented the design, modeling and experimental validation of a low-cost underwater quadruped robot for attitude control. The open-source hardware design relies on off-the-shelf servo motors, electronics and sealing components combined with POM-machined casings, providing a reproducible and cost-effective waterproofing solution for articulated underwater robots. A simplified dynamics model is obtained from floating base kinematics and drag-based swimming and used to design a closed-loop SO(3) controller, which was deployed on the physical platform for reorientation in roll, pitch and coupled maneuvers.

\subsubsection*{Acknowledgements.}
This work was supported by the Research Council of Norway as part of the Norwegian Centre for Embodied AI (357451). 

\bibliographystyle{ieeetr}
\bibliography{BIB/main}

\end{document}